\documentclass{article}
\usepackage[utf8]{inputenc}
\usepackage[preprint,nonatbib]{neurips}
\usepackage{bm}
\usepackage{graphicx}
\usepackage{hyperref}
\usepackage{enumitem}
\usepackage{caption}
\usepackage{subcaption}
\usepackage{hyperref}
\usepackage{nicefrac, xfrac}
\author{Yukun Yang\\
Institute of Theoretical Computer Science\\
  Graz University of Technology\\
  \texttt{yukun.yang@tugraz.at} \\}
\usepackage{lineno, color}
\usepackage[utf8]{inputenc} 
\usepackage[T1]{fontenc}    
\usepackage{hyperref}       
\usepackage{url}            
\usepackage{booktabs}       
\usepackage{amsfonts}       
\usepackage{nicefrac}       
\usepackage{microtype}      
\usepackage{xcolor}         
\usepackage{amsmath}
\usepackage{algorithm} 
\usepackage{amsthm}
\usepackage{dsfont}
\newtheorem{theorem}{Theorem}
\usepackage{algpseudocode} 
\date{February 2023}
\usepackage[
backend=biber,
natbib=true,
style=authoryear,
sorting=nty
]{biblatex}

\usepackage{tabularx}
\title{A theory for the sparsity emerged in the Forward Forward algorithm}

\begin{document}

\maketitle
\begin{abstract}
This report explores the theory that explains the high sparsity phenomenon \citep{tosato2023emergent} observed in the forward-forward algorithm \citep{hinton2022forward}. The two theorems proposed predict the sparsity changes of a single data point's activation in two cases:
    Theorem \ref{theorem:1}: Decrease the goodness of the whole batch.
    Theorem \ref{theorem:2}: Apply the complete forward forward algorithm to decrease the goodness for negative data and increase the goodness for positive data.
The theory aligns well with the experiments tested on the MNIST dataset.
\end{abstract}



\section{Definition}
\subsection{Linear layer}
Let $\bm{x} \in \mathbb{R}^m$ be the input vector to the fully-connected layer with $\mathbf{W} \in \mathbb{R}^{n \times m}$ the weight matrix. The output activation of the fully-connected layer with the ReLU activation is given by:

\begin{equation}
    \bm{h} = \text{ReLU}(\mathbf{W}\bm{x}) \in \mathbb{R}^n,{\rm ~~~where~~~} \text{ReLU}(\bm{v})[p] =\left\{\begin{aligned}
    &\bm{v}[p], \text{~if~}\bm{v}[p]>0\\
    &0, \text{~otherwise}
    \end{aligned}\right.,
\end{equation}

where $\bm{v}[p]$ is the $p$th element of the vector $\bm{v}$. Further, we define a mask function $m$ indicating if each dimension of $\bm{h}$ is positive or been masked out by ReLU to zero:

\begin{equation}
    \bm{m}\in \mathbb{R}^n\text{~~~,~~~}\bm{m}[p]=\left\{\begin{aligned}
    &1, \text{~if~}\bm{h}[p]>0\\
    &0, \text{~otherwise}
    \end{aligned}\right..
\end{equation}

This proof only discusses the ReLU activation.

\subsection{Goodness}
For a batch of input vectors $\mathcal{X}=\{\bm{x}_1, \bm{x}_2, \cdots, \bm{x}_N\}$ with batch size $=N$, the corresponding outputs are defined as $\mathcal{H}=\{\bm{h}_1, \bm{h}_2, \cdots, \bm{h}_N\}$, where $\bm{h}_i =\text{ReLU}(\mathbf{W}\bm{x}_i)$. 
We define the goodness as the summation of all hidden activations' $l_2$ norm minus a fixed threshold $\theta$, i.e.:

\begin{equation}
    G(\mathcal{H}) =\sum_{i=1}^N\|\bm{h}_i\|_2^2 - \theta.
\end{equation}

\subsection{Sparsity}
The sparsity $S$ of an activation vector $\bm{h}$ is defined as \citep{hoyer2004non}:

\begin{equation}
    S(\bm{h}) = \frac{\sqrt{n} - \frac{\|\bm{h}\|_1}{\|\bm{h}\|_2}}{\sqrt{n} - 1}.
\end{equation}

The sparsity measure $S(\bm{h})\in[0,1]$. A higher value of $S(\bm{h})$ indicates a sparser vector $\bm{h}$.

\subsection{Forward forward algorithm}

We define two group of data: A batch of positive data  $\mathcal{X}_+=\{\bm{x}_{+1}, \bm{x}_{+2}, \cdots, \bm{x}_{+N}\}$ with batch size $=N$, the corresponding outputs are defined as $\mathcal{H}_+=\{\bm{h}_{+1}, \bm{h}_{+2}, \cdots, \bm{h}_{+N}\}$, where $\bm{h}_{+i} =\text{ReLU}(\mathbf{W}\bm{x}_{+i})$; and a batch of negative data $\mathcal{X}_-=\{\bm{x}_{-1}, \bm{x}_{-2}, \cdots, \bm{x}_{-N}\}$ with batch size $=N$, the corresponding outputs are defined as $\mathcal{H}_-=\{\bm{h}_{-1}, \bm{h}_{-2}, \cdots, \bm{h}_{-N}\}$, where $\bm{h}_{-i} =\text{ReLU}(\mathbf{W}\bm{x}_{-i})$.

Then, define loss $L$ as:

\begin{equation}
    L=G(\mathcal{H}_-) - G({\mathcal{H}_+})
\end{equation}

Minimizing $L$ is equivalent to increase positive data's goodness while decrease negative data's goodness. In each iteration, we perform the stochastic gradient descent to train the weights $W$ with a batch of positive data and a batch of negative data.

\section{Results}

\textbf{Assumptions:}

\begin{enumerate}
    \item The width of the linear layer $n\rightarrow +\infty$
    \label{list:1}
    \item Learning rate $\eta=\mathcal{O}(\frac{1}{n})$
    \label{list:2}
    \item The pre-activation input $\mathbf{W}\bm{x}$ does not have zero element: $\forall \bm{x}_i\in\mathbf{X}, \forall p, (\mathbf{W}\bm{x}_i)[p]\neq 0$. (Note: what I assumed here is neither the input $\bm{x}$ nor the output $\bm{h}$ which usually has many zero elements. The weighted sum $\mathbf{W}\bm{x}$ can rarely equals to exact zero.)
    \label{list:3}
    \item All inputs $\bm{x}_i\in\mathbf{X}$ are normalized to have a unit length: $\forall i,~\|\bm{x}_i\|_2=1$
    \label{list:4}
    \item Weights are initialized by a Gaussian distribution $\mathbf{W}\sim\mathcal{N}(0,\frac{2}{n})$. The variance $\sigma=\frac{2}{n}$ is picked following the Kaiming initialization \citep{he2015delving}, which guarantees the output activation $\bm{h}$ has the same scale of $l_2$ norm as compared to the input $\bm{x}$.
    \label{list:5}
    \item All inputs and all layers at initialization are iid, i.e. independent, identically distributed, and thus, so are the weights and gradients.
    \label{list:6}
    \item The non-linear activation function is ReLU.
    \label{list:7}
\end{enumerate}

It is not always guaranteed that all samples would becomes sparser in the forward forward algorithm. After each weight adjustment, there are always samples' sparsity that are increased, unchanged, or decreased. This depends on the property of each data point. A property that, if the majority of data points satisfy, leads to a sparser activation for all in general.

This proof begins with a simpler scenario: We wish to know under what property of a data point that \textbf{decreasing} the goodness of the whole batch would \textbf{increase} the data point's sparsity. This gives us Theorem \ref{theorem:1}.

~

\begin{theorem}
    If a data point $\bm{x}_i$ with its activation $\bm{h}_i$ satisfies:
    \begin{equation}
        \frac{\|\bm{h}_i\|_1}{\|\bm{h}_i\|_2}<\frac{\sum_{k=1}^N \|\bm{h}_{k}\odot \bm{m}_i\|_1   cos(\bm{x}_k,\bm{x}_i)}{\sum_{k=1}^N\|\bm{h}_{k}\odot \bm{m}_i\|_2    cos(\bm{h}_i, \bm{h}_{k}\odot \bm{m}_i)    cos(\bm{x}_k,\bm{x}_i)},
        \label{eq:property}
    \end{equation}
    then decreasing the goodness $G(\mathcal{H})$ following gradient descent would increase $\bm{h}_i$'s sparsity $S(\bm{h}_i)$.
    \label{theorem:1}
\end{theorem}

~

In above, $\odot$ is the elementwise product. Importantly to know, the reverse order of this theorem is also true. If decreasing the goodness $G(\mathcal{H})$ following gradient descent would increase $\bm{h}_i$'s sparsity $S(\bm{h}_i)$, then increase the goodness $G(\mathcal{H})$ following gradient descent would decrease $S(\bm{h}_i)$. We cannot always increase sparsity no matter if we decrease or increase the goodness. The phenomenon that both positive and negative samples have their activations sparsity increased in the forward forward algorithm cannot be studied separately. It must be studied as a co-effect of the co-existence of both groups of data samples.


Therefore, under the same assumptions, we derived the following Theorem for the complete setting of the forward forward algorithm, where there exist both positive data $\mathcal{X}_+$ and negative data $\mathcal{X}_-$, and their corresponding activations $\mathcal{H}_+$ and  $\mathcal{H}_-$:

~

\begin{theorem}
    If a data point $\bm{x}_i$ (No matter if $\bm{x}_i\in \mathcal{X}_-$ or $\bm{x}_i\in \mathcal{X}_+$) with its activation $\bm{h}_i$ satisfies :
    \begin{equation}
    \|\bm{h}_i\|_2(A_+-A_-) < \|\bm{h}_i\|_1(B_+-B_-)
    \label{eq:property_2}
    \end{equation}
    where
    \begin{equation}
        \begin{aligned}
            A_+ & = \sum_{k=1}^N \|\bm{h}_{+k} \odot \bm{m}_i\|_1 cos(\bm{x}_{+k},\bm{x}_i)\\
            A_- & = \sum_{k=1}^N \|\bm{h}_{-k} \odot \bm{m}_i\|_1 cos(\bm{x}_{-k},\bm{x}_i)\\
            B_+ & = \sum_{k=1}^N\|\bm{h}_{+k}\odot \bm{m}_i\|_2 cos(\bm{h}_i, \bm{h}_{+k}\odot \bm{m}_i) cos(\bm{x}_{+k},\bm{x}_i)\\
            B_- & = \sum_{k=1}^N\|\bm{h}_{-k}\odot \bm{m}_i\|_2 cos(\bm{h}_i, \bm{h}_{-k}\odot \bm{m}_i) cos(\bm{x}_{-k},\bm{x}_i),
        \end{aligned}
    \end{equation}
    then decreasing the loss $L=G(\mathcal{H}_-) - G({\mathcal{H}_+})$ following gradient descent would increase $\bm{h}_i$'s sparsity $S(\bm{h}_i)$.
    \label{theorem:2}
\end{theorem}

\section{Discussion}

The key findings are (Please refer to the experimental results in the later sections for detail):

\begin{itemize}
    \item Most of the data points satisfy Theorem \ref{theorem:1} at the beginning of training: Decrease the goodness would increase their sparsity, and increase the goodness would decrease their sparsity.
    \item When consider the co-effects of both positive and negative data, we observed that there are usually more than half data points satisfy Theorem \ref{theorem:2} and are becoming sparser in training. This result holds for both positive and negative data.
\end{itemize}

To dig out more insights from them, I am considering the following directions like:

\begin{itemize}
    \item Interpret the intuitive meaning of the theorems, and explain why they are (Especially Theorem \ref{theorem:2}) satisfied by most of the data points in a dataset like MNIST and CIFAR10 in general.
    \item Consider principles of designing the negative data based on Theorem \ref{theorem:2}.
    \item Simplify the theorems by bringing in more details from experiments. For example, consider what if negative samples are generated by substituting positive samples' labels by randomly generated wrong labels, so that $\bm{x}_-i$ differs from $\bm{x}_+i$ by just two pixels.
    \item Modularize the theorems to be more general, such that the effects on the sparsity after changing the goodness function, the normalization method, and others can be better predicted/explained.
    \item Consider the extended versions like on recurrent neural networks, and on spiking neurons with temporal dynamics to guide neuromorphic applications.
\end{itemize}

\newpage
\section{Proof}

~

\textbf{Sparsity comparison}

Naming $\bm{h}_i$ and $\bm{h}_i'$ as the activations before and after an weight update respectively, our strategy is to compare the value of the sparsity measures $S(\bm{h}_i)$ and $S(\bm{h}_i')$:

\begin{equation}
    \text{(Left~hand~side)~~~}\frac{\sqrt{n} - \frac{\|\bm{h}_i\|_1}{\|\bm{h}_i\|_2}}{\sqrt{n} - 1} \text{~~~v.s.~~~} \frac{\sqrt{n} - \frac{\|\bm{h}_i'\|_1}{\|\bm{h}_i'\|_2}}{\sqrt{n} - 1}\text{~~~(Right~hand~side)}.
    \label{eq:compare}
\end{equation}

We aim to demonstrate that the sparsity prior to the update, represented by the left-hand side, is less than or equal to the sparsity following the update, represented by the right-hand side, when $\bm{h}_i$ fulfills the condition specified in both theorems. We start by simplifying this comparison to its equivalent forms:

\begin{equation}
\begin{aligned}
    \Longleftrightarrow \sqrt{n} - \frac{\|\bm{h}_i\|_1}{\|\bm{h}_i\|_2} &\text{~~~v.s.~~~} \sqrt{n} - \frac{\|\bm{h}_i'\|_1}{\|\bm{h}_i'\|_2}, \\
    \Longleftrightarrow \frac{\|\bm{h}_i'\|_1}{\|\bm{h}_i'\|_2} &\text{~~~v.s.~~~} \frac{\|\bm{h}_i\|_1}{\|\bm{h}_i\|_2}, \\
    \Longleftrightarrow \|\bm{h}_i'\|_1\|\bm{h}_i\|_2 &\text{~~~v.s.~~~}\|\bm{h}_i\|_1\|\bm{h}_i'\|_2.
\end{aligned}
\end{equation}

Further, when we replace $|\bm{h}_i'|_1$ with its delta changes $\|\bm{h}_i\|_1+\Delta_{l_1}$, and likewise substitute $\|\bm{h}_i'\|_2$ with $\|\bm{h}_i\|_2+\Delta_{l_2}$, we obtain the comparison as:

\begin{equation}
\begin{aligned}
    \Longleftrightarrow (\|\bm{h}_i\|_1+\Delta_{l_1})\|\bm{h}_i\|_2 &\text{~~~v.s.~~~}\|\bm{h}_i\|_1(\|\bm{h}_i\|_2+\Delta_{l_2}),\\
    \Longleftrightarrow \Delta_{l_1}\|\bm{h}_i\|_2 &\text{~~~v.s.~~~}\|\bm{h}_i\|_1\Delta_{l_2}.
    \label{eq:compare_trans}
\end{aligned}
\end{equation}

Such a comparison requires an evaluation on the changes of $\bm{h}_i$'s $l_1$ and $l_2$ norms: $\Delta{l_1}$ and $\Delta{l_2}$. Next, we will exam how the update on weights $\mathbf{W}$ affect these two values.

~

\subsection{Proof of Theorem \ref{theorem:1}}

~

\textbf{Weight update:}

Recalling the goodness $G(\mathcal{H}) =\sum_{k=1}^N\|\bm{h}_k\|_2^2-\theta$, we have the partial derivative of $G$ w.r.t the weight $\mathbf{W}$ as:

\begin{equation}
    \frac{\partial G}{\partial\mathbf{W}}= \sum_{k=1}^N \frac{\partial G}{\partial \bm{h}_k} \frac{\partial \bm{h}_k}{\partial\mathbf{W}}= \sum_{k=1}^N 2 (\bm{h}_k \odot \bm{m}_k)  \bm{x}_k^T .
    \label{eq:dW}
\end{equation}

The term $\bm{m}_k$ comes from the derivative of $\text{ReLU}$ function, which is defined as:

\begin{equation}
    \bm{m}_k[p]=\left\{\begin{aligned}
    &1, \text{~if~}\bm{h}_k[p]>0\\
    &0, \text{~otherwise}
    \end{aligned}\right..
\end{equation}

We notice that:

\begin{equation}
    (\bm{h}_k \odot \bm{m}_k)[p]=\left\{\begin{aligned}
    &\bm{h}_k[p], \text{~if~}\bm{h}_k[p]>0\\
    &0, \text{~otherwise}
    \end{aligned}\right. \text{~~~equals~~to~~~} \bm{h}_k,
\end{equation}

so $\bm{m}_k$ can be absorbed into $\bm{h}_k$, and (\ref{eq:dW}) can be further simplified as:

\begin{equation}
    \frac{\partial G}{\partial\mathbf{W}}= \sum_{k=1}^N 2 \bm{h}_k   \bm{x}_k^T.
    \label{eq:dW}
\end{equation}

Applying gradient descent on W with the step size $\eta$, we have the weight change $\Delta\mathbf{W}$ as:

\begin{equation}
    \Delta\mathbf{W} = -2\eta \sum_{k=1}^N \bm{h}_k   \bm{x}_k^T.
\end{equation}

Therefore, the activation $\bm{h}_i$ and its updated activation $\bm{h}_i'$ are:

\begin{equation}
    \bm{h}_i= \text{ReLU}\left(\mathbf{W}\bm{x}_i\right),~~~\text{and~~~}\bm{h}_i'= \text{ReLU}\left[\left(\mathbf{W}-2\eta\sum_{k=1}^N \bm{h}_k \bm{x}_k^T\right)\bm{x}_i\right].
\end{equation}

Next, we compare the above two vectors, and measure the changes of its $l_1$ and $l_2$ norms separately.

~

\textbf{Change of the }$\bm{l_1}$\textbf{ norm:}

\begin{equation}
    \Delta_{l_1} = \|\bm{h}_i'\|_1 - \|\bm{h}_i\|_1.
\end{equation}

With ReLU activation, all elements in both $\bm{h}_i$ and $\bm{h}_i'$ are non-negative, so $\forall p, \bm{h}_i[p]\geq0$ and $\bm{h}_i'[p]\geq0$, and the $l_1$ norm is simply the summation of all elements:

\begin{equation}
\begin{aligned}
    \Delta_{l_1} &= \|\bm{h}_i'\|_1 - \|\bm{h}_i\|_1 = \sum_{p=1}^n \bm{h}_i'[p] - \sum_{p=1}^n \bm{h}_i[p],\\
    &=\mathds{1}^T(\bm{h}_i'-\bm{h}_i).
\end{aligned}
\label{eq:l1_relu}
\end{equation}

In above, $\mathds{1}$ is an all-one vector with length $n$. The scalar product between $\mathds{1}$ and any other vector $\bm{h}$ with the same length results in a summation of all terms of $\bm{h}$:

\begin{equation}
  \mathds{1}^T\bm{h}=\sum_{p=1}^nh[p].
\end{equation}

Bringing in the value of both $\bm{h}_i'$ and $\bm{h}_i$ into (\ref{eq:l1_relu}), we get:

\begin{equation}
    \Delta_{l_1} =\mathds{1}^T\left[\text{ReLU}\left((\mathbf{W}-2\eta\sum_{k=1}^N \bm{h}_k \bm{x}_k^T)\bm{x}_i\right) - \text{ReLU}\left(\mathbf{W}\bm{x}_i\right)\right].
\end{equation}

Next, we will simplify the above equation by proving the sign of each element in $\text{ReLU}$ before and after the weight update is the same.

With assumption \ref{list:2}, we notice the changes $\Delta$ inside the ReLU before and after the weight update is an infinitesimal vector when $n\rightarrow+\infty$:

\begin{equation}
    \Delta = -2\eta\sum_{k=1}^N \bm{h}_k \bm{x}_k^T \bm{x}_i=\mathcal{O}(\frac{1}{n})\sum_{k=1}^N \bm{h}_k\bm{x}_k^T \bm{x}_i,
\end{equation}

From assumption \ref{list:4}, we have  $\bm{x}_k^T \bm{x}_i=cos(\bm{x}_k,\bm{x}_i)\leq 1$, which does not enlarge the magnitude of the resulting vector. Since the batch size $N$ is a constant, ($n>>N$), $\sum_{k=1}^N \bm{h}_k$ is among the same magnitude as any $\bm{h}_k$.

As compared to $\mathbf{W}\bm{x}_i$, we can always have a $\Delta$ that has a smaller absolute value on every element by increasing $n$. Therefore, with assumption \ref{list:1}, we can guarantee: For $\bm{h}_i'$ and $\bm{h}_i$, the signs of all elements in ReLU does not change after a weight update:

\begin{equation}
    \text{sign}\left[\left(\mathbf{W}-2\eta\sum_{k=1}^N \bm{h}_k \bm{x}_k^T\right)\bm{x}_i\right] = \text{sign} \left(\mathbf{W}\bm{x}_i\right),
\end{equation}

With this property, the subtraction between two $\text{ReLU}$ terms can be discussed separately based on the sign inside $\text{ReLU}$.

For the $p$-th element, if $(\mathbf{W}\bm{x}_i[p])>0$, we can remove ReLU in (\ref{eq:l1_relu}) and have the change $\bm{h}_i'[p]-\bm{h}_i[p]$ equals to $\Delta[p]$. Whereas if $(\mathbf{W}\bm{x}_i)[p]<0$, then we have both $\bm{h}_i[p]=\bm{h}_i'[p]=0$. So the update has no change on $\bm{h}_i[p]$.  As a conclusion, we get:

\begin{equation}
    \bm{h}_i'[p]-\bm{h}_i[p] = \left\{\begin{aligned}
    &\Delta[p], \text{~if~} (\mathbf{W}\bm{x}_i)[p]>0\\
    &0, \text{~if~} (\mathbf{W}\bm{x}_i)[p]<0
    \end{aligned}\right..
\end{equation}

Here we intentionally omit the situation when $(\mathbf{W}\bm{x}_i[p])=0$ by assuming this is not exist (assumption \ref{list:3}). This assumption is generally true for the weighted sum of an input's all dimension can hardly be zero if the input $\bm{x}_i$ is not an all-zero vector. We will further verify this through experiments.

Such an effect can be perfectly captured by masking out the negative part of $\Delta$ by $\bm{m}_i$:

\begin{equation}
    \bm{h}_i'-\bm{h}_i = -2\eta\sum_{k=1}^N \bm{h}_k \bm{x}_k^T\bm{x}_i\odot \bm{m}_i.
    \label{eq:h_change}
\end{equation}

Bring the above equation into (\ref{eq:l1_relu}), we have:

\begin{equation}
\begin{aligned}
    \Delta_{l_1} &= \mathds{1}^T(-2\eta\sum_{k=1}^N \bm{h}_k \bm{x}_k^T\bm{x}_i\odot \bm{m}_i),\\
    &=-2\eta\sum_{k=1}^N cos(\bm{x}_k,\bm{x}_i) \mathds{1}^T(\bm{h}_k \odot \bm{m}_i).
\end{aligned}
\label{eq:l1_mask}
\end{equation}

Since all elements in $(\bm{h}_k\odot \bm{m}_i)$ are non-negative with the ReLU activation, the summation of all elements is equivalent to calculating its $l_1$ norm. We arrive at our final form of the $\Delta_{l_1}$ as:  

\begin{equation}
    \Delta_{l_1}=-2\eta\sum_{k=1}^N cos(\bm{x}_k,\bm{x}_i) \| \bm{h}_k \odot \bm{m}_i\|_1.
\label{eq:l1_mask}
\end{equation}

The above equation is an accurate measure of the changes on $\bm{h}_i$'s $l_1$ norm before and after a weight update.

~

\textbf{Change of the }$\bm{l_2}$\textbf{ norm:}

\begin{equation}
    \Delta_{l_2} = \|\bm{h}_i'\|_2 - \|\bm{h}_i\|_2.
    \label{eq:l2_change}
\end{equation}

Performing multi-dimension Taylor expansion, we get:

\begin{equation}
    \|\bm{h}_i'\|_2 = \|\bm{h}_i\|_2 + (\bm{h}_i'-\bm{h}_i)^T  \mathcal{D}(\|\bm{h}_i\|_2) + \frac{1}{2!}(\bm{h}_i'-\bm{h}_i)^T  \mathcal{D}^2(\|\bm{h}_i\|_2)  (\bm{h}_i'-\bm{h}_i)+\cdots,
    \label{eq:taylor}
\end{equation}

where $\mathcal{D}(\|\bm{h}_i\|_2)$ is the gradient of the $l_2$ norm at $\bm{h}_i$, and $\mathcal{D}^2(\|\bm{h}_i\|_2)$ is the Hessian matrix. The gradient is:

\begin{equation}
    \mathcal{D}(\|\bm{h}_i\|_2) = (\frac{\partial \|\bm{h}_i\|_2}{\partial \bm{h}_i[1]}, \frac{\partial \|\bm{h}_i\|_2}{\partial \bm{h}_i[2]},\cdots, \frac{\partial \|\bm{h}_i\|_2}{\partial \bm{h}_i[n]})^T.
    \label{eq:l2_all}
\end{equation}

For each term in the above equation, we have:

\begin{equation}
    \frac{\partial \|\bm{h}_i\|_2}{\partial \bm{h}_i[p]} = \frac{\partial \sqrt{\sum_{q=1}^n \bm{h}_i[q]^2}}{\partial \bm{h}_i[p]}= \frac{2\bm{h}_i[p]}{2\sqrt{\sum_{q=1}^n \bm{h}_i[q]^2}} = \frac{\bm{h}_i[p]}{\|\bm{h}_i\|_2}.
    \label{eq:l2_dim}
\end{equation}

Bring (\ref{eq:l2_dim}) back into (\ref{eq:l2_all}), we have the gradient as:

\begin{equation}
    \mathcal{D}(\|\bm{h}_i\|_2) = \frac{1}{\|\bm{h}_i\|_2} \bm{h}_i.
    \label{eq:grad}
\end{equation}

Represent the higher terms by their big-O notation:

\begin{equation}
    \frac{1}{2!}(\bm{h}_i'-\bm{h}_i)^T  \mathcal{D}^2(\|\bm{h}_i\|_2)  (\bm{h}_i'-\bm{h}_i)+\cdots = \mathcal{O}(\|\bm{h}_i'-\bm{h}_i\|_2^2).
    \label{eq:bigO}
\end{equation}

Bring (\ref{eq:h_change}) into the above equation, we get:

\begin{equation}
    \mathcal{O}(\|\bm{h}_i'-\bm{h}_i\|_2^2) = \mathcal{O}(\|-2\eta\sum_{k=1}^N \bm{h}_k \bm{x}_k^T\bm{x}_i\odot \bm{m}_i\|_2^2).
\end{equation}

In the above equation, we know $\eta=\mathcal{O}(\frac{1}{n})$ from assumption \ref{list:2}; and we know $\bm{x}_k^T\bm{x}_i=cos(\bm{x}_k,\bm{x}_i)<1$ from assumption \ref{list:4}. Based on assumption \ref{list:5}, the output $\bm{h}$ and the input $\bm{x}$ shall be in the same magnitude. Considering the dataset's size $N$ as a constant, we have $\|\sum_{k=1}^N \bm{h}_k\odot \bm{m}_i\|_2^2=\mathcal{O}(1)$. Combining them all, we have:

\begin{equation}
\mathcal{O}(\|\bm{h}_i'-\bm{h}_i\|_2^2) = \mathcal{O}(\frac{1}{n^2}).
\label{eq:bigO_r}
\end{equation}

Bringing (\ref{eq:taylor}), (\ref{eq:grad}), (\ref{eq:bigO}) and (\ref{eq:bigO_r}) into (\ref{eq:l2_change}), we get:

\begin{equation}
    \Delta_{l_2} = (\bm{h}_i'-\bm{h}_i)^T\frac{1}{\|\bm{h}_i\|_2} \bm{h}_i + \mathcal{O}(\frac{1}{n^2}).
    \label{eq:l2_taylor}
\end{equation}

Further bring (\ref{eq:h_change}) into the above equation to substitute $(\bm{h}_i'-\bm{h}_i)$:

\begin{equation}
\begin{aligned}
    \Delta_{l_2} &= \left(-2\eta\sum_{k=1}^N \bm{h}_k \bm{x}_k^T\bm{x}_i\odot \bm{m}_i\right)^T \frac{1}{\|\bm{h}_i\|_2} \bm{h}_i + \mathcal{O}(\frac{1}{n^2}),\\
    &=-2\eta\frac{1}{\|\bm{h}_i\|_2}\left(\sum_{k=1}^N (\bm{h}_k\odot \bm{m}_i) cos(\bm{x}_k,\bm{x}_i)\right)^T \bm{h}_i  + \mathcal{O}(\frac{1}{n^2}).
\end{aligned}
\end{equation}

To further simplify the above equation, we substitute the scalar product between the two vectors $(\bm{h}_k\odot \bm{m}_i)$ and $\bm{h}_i$ as their cosine similarity multiplies their $l_2$ norms:

\begin{equation}
\begin{aligned}
    \Delta_{l_2} &=-2\eta\frac{1}{\|\bm{h}_i\|_2}\sum_{k=1}^N \|\bm{h}_k\odot \bm{m}_i\|_2\|\bm{h}_i\|_2 cos(\bm{h}_k\odot \bm{m}_i, \bm{h}_i) cos(\bm{x}_k,\bm{x}_i)   + \mathcal{O}(\frac{1}{n^2}),\\
    &=-2\eta\sum_{k=1}^N \|\bm{h}_k\odot \bm{m}_i\|_2 cos(\bm{h}_k\odot \bm{m}_i, \bm{h}_i) cos(\bm{x}_k,\bm{x}_i)   + \mathcal{O}(\frac{1}{n^2}).
\end{aligned}
\label{eq:dl2_r}
\end{equation}

Now we have the change on both $l_1$ and $l_2$ norms. Next, we will bring them back to the sparsity comparison.

~

\textbf{Final step}

Recall comparing the value of the sparsity measures $S(\bm{h}_i)$ and $S(\bm{h}_i')$:

\begin{equation}
    \text{(Left~hand~side)~~~}\frac{\sqrt{n} - \frac{\|\bm{h}_i\|_1}{\|\bm{h}_i\|_2}}{\sqrt{n} - 1} \text{~~~v.s.~~~} \frac{\sqrt{n} - \frac{\|\bm{h}_i'\|_1}{\|\bm{h}_i'\|_2}}{\sqrt{n} - 1}\text{~~~(Right~hand~side)},
\end{equation}

is equivalent to compare

\begin{equation}
    \text{(Left~hand~side)~~~} \Delta_{l_1}\|\bm{h}_i\|_2 \text{~~~v.s.~~~}\|\bm{h}_i\|_1\Delta_{l_2}\text{~~~(Right~hand~side)}.
    \label{eq:compare_trans}
\end{equation}

Bringing the calculated $\Delta_{l_1}$ in (\ref{eq:l1_mask}) and $\Delta_{l_1}$ in (\ref{eq:dl2_r}) into the above comparison, we get:

\begin{equation}
\begin{aligned}
    -2\eta\sum_{k=1}^N &cos(\bm{x}_k,\bm{x}_i) \| \bm{h}_k \odot \bm{m}_i\|_1\|\bm{h}_i\|_2\text{~~~v.s.~~~}\\
    &\|\bm{h}_i\|_1\left[-2\eta\sum_{k=1}^N \|\bm{h}_k\odot \bm{m}_i\|_2 cos(\bm{h}_k\odot \bm{m}_i, \bm{h}_i) cos(\bm{x}_k,\bm{x}_i)   + \mathcal{O}(\frac{1}{n^2})\right].\\
\end{aligned}
\end{equation}

Next, we divide both sides by $(-2\eta)$. Since $(-2\eta)<0$, we need to swap the two sides after dividing:

\begin{equation}
    \|\bm{h}_i\|_1\left[\sum_{k=1}^N \|\bm{h}_{k\bm{m}_i}\|_2 cos(\bm{h}_{k\bm{m}_i}, \bm{h}_i) cos(\bm{x}_k,\bm{x}_i)   + \mathcal{O}(\frac{1}{n})\right]\text{~~v.s.~~}
    \sum_{k=1}^N cos(\bm{x}_k,\bm{x}_i) \| \bm{h}_{k\bm{m}_i}\|_1\|\bm{h}_i\|_2,
\end{equation}

where $\bm{h}_{k\bm{m}_i}=\bm{h}_k\odot \bm{m}_i$ for short. Since $\eta=\mathcal{O}(\frac{1}{n})$, dividing $\mathcal{O}(\frac{1}{n^2})$ by $-2\eta$ results in $\mathcal{O}(\frac{1}{n})$. Reformulate the above equation gives us the final form:

\begin{equation}
    \frac{\|\bm{h}_i\|_1}{\|\bm{h}_i\|_2}\text{~~v.s.~~} \frac{\sum_{k=1}^N cos(\bm{x}_k,\bm{x}_i) \| \bm{h}_{k\bm{m}_i}\|_1}{\sum_{k=1}^N \|\bm{h}_{k\bm{m}_i}\|_2 cos(\bm{h}_{k\bm{m}_i}, \bm{h}_i) cos(\bm{x}_k,\bm{x}_i)   + \mathcal{O}(\frac{1}{n})}.
\end{equation}

We can omit $\mathcal{O}(\frac{1}{n})$ as $n\rightarrow+\infty$. 

In conclusion, an activation $\bm{h}_i$ becomes sparser after weight update:

\begin{equation}
    \text{(Left~hand~side)~~~}\frac{\sqrt{n} - \frac{\|\bm{h}_i\|_1}{\|\bm{h}_i\|_2}}{\sqrt{n} - 1} < \frac{\sqrt{n} - \frac{\|\bm{h}_i'\|_1}{\|\bm{h}_i'\|_2}}{\sqrt{n} - 1}\text{~~~(Right~hand~side)},
\end{equation}

is equivalent to requiring:

\begin{equation}
    \frac{\|\bm{h}_i\|_1}{\|\bm{h}_i\|_2} < \frac{\sum_{k=1}^N cos(\bm{x}_k,\bm{x}_i) \| \bm{h}_{k\bm{m}_i}\|_1}{\sum_{k=1}^N \|\bm{h}_{k\bm{m}_i}\|_2 cos(\bm{h}_{k\bm{m}_i}, \bm{h}_i) cos(\bm{x}_k,\bm{x}_i)}.
\end{equation}

\begin{flushright}
\textbf{[Q.E.D.]}
\end{flushright}

\newpage
\subsection{Proof of Theorem \ref{theorem:2}}

\textbf{Weight update:}

Recalling the goodness $G(\mathcal{H}) =\sum_{k=1}^N\|\bm{h}_k\|_2^2-\theta$, and the loss $L=G(\mathcal{H}_-) - G({\mathcal{H}_+})$ we have the partial derivative of $L$ w.r.t the weight $\mathbf{W}$ as:

\begin{equation}
    \frac{\partial L}{\partial\mathbf{W}}= \sum_{k=1}^N \frac{\partial L}{\partial \bm{h}_{-k}} \frac{\partial \bm{h}_{-k}}{\partial\mathbf{W}} - \sum_{k=1}^N \frac{\partial L}{\partial \bm{h}_{+k}} \frac{\partial \bm{h}_{+k}}{\partial\mathbf{W}}= \sum_{k=1}^N 2 (\bm{h}_{-k} \odot \bm{m}_{-k})  \bm{x}_k^T - \sum_{k=1}^N 2 (\bm{h}_{+k} \odot \bm{m}_{+k})  \bm{x}_k^T .
    \label{eq:dW}
\end{equation}

The term $\bm{m}_k$ comes from the derivative of $\text{ReLU}$ function, which is defined as:

\begin{equation}
    \bm{m}_k[p]=\left\{\begin{aligned}
    &1, \text{~if~}\bm{h}_k[p]>0\\
    &0, \text{~otherwise}
    \end{aligned}\right..
\end{equation}

We notice that:

\begin{equation}
    (\bm{h}_k \odot \bm{m}_k)[p]=\left\{\begin{aligned}
    &\bm{h}_k[p], \text{~if~}\bm{h}_k[p]>0\\
    &0, \text{~otherwise}
    \end{aligned}\right. \text{~~~equals~~to~~~} \bm{h}_k,
\end{equation}

so $\bm{m}_k$ can be absorbed into $\bm{h}_k$, and (\ref{eq:dW}) can be further simplified as:

\begin{equation}
    \frac{\partial G}{\partial\mathbf{W}}= \sum_{k=1}^N 2 \bm{h}_{-k} \bm{x}_{-k}^T - \sum_{k=1}^N 2 \bm{h}_{+k} \bm{x}_{+k}^T.
    \label{eq:dW}
\end{equation}

Applying gradient descent on W with the step size $\eta$, we have the weight change $\Delta\mathbf{W}$ as:

\begin{equation}
    \Delta\mathbf{W} = -2\eta \sum_{k=1}^N \left(\bm{h}_{-k} \bm{x}_{-k}^T - \bm{h}_{+k} \bm{x}_{+k}^T\right).
\end{equation}

Therefore, the activation $\bm{h}_i$ and its updated activation $\bm{h}_i'$ are:

\begin{equation}
    \bm{h}_i= \text{ReLU}\left(\mathbf{W}\bm{x}_i\right),~~~\text{and~~~}\bm{h}_i'= \text{ReLU}\left[\left(\mathbf{W}-2\eta\sum_{k=1}^N \left(\bm{h}_{-k} \bm{x}_{-k}^T - \bm{h}_{+k} \bm{x}_{+k}^T\right)\right)\bm{x}_i\right].
\end{equation}

Next, we compare the above two vectors, and measure the changes of its $l_1$ and $l_2$ norms separately.

~

\textbf{Change of the }$\bm{l_1}$\textbf{ norm:}

Following the same logic as (\ref{eq:h_change}) in the previous section, we can get:

\begin{equation}
    \bm{h}_i'-\bm{h}_i = -2\eta\sum_{k=1}^N \left(\bm{h}_{-k} \bm{x}_{-k}^T\bm{x}_i\odot \bm{m}_i - \bm{h}_{+k} \bm{x}_{+k}^T\bm{x}_i\odot \bm{m}_i\right).
    \label{eq:h_change_2}
\end{equation}

Bring this into (\ref{eq:l1_relu}), we get:

\begin{equation}
\begin{aligned}
    \Delta_{l_1} &=\mathds{1}^T(\bm{h}_i'-\bm{h}_i),\\
    &= \mathds{1}^T(-2\eta\sum_{k=1}^N \left(\bm{h}_{-k} \bm{x}_{-k}^T\bm{x}_i\odot \bm{m}_i - \bm{h}_{+k} \bm{x}_{+k}^T\bm{x}_i\odot \bm{m}_i\right)), \\
    &= -2\eta \sum_{k=1}^N\left( \|\bm{h}_{-k}\odot \bm{m}_i\|_1 cos(\bm{x}_{-k}, \bm{x}_i)-\|\bm{h}_{+k}\odot \bm{m}_i\|_1 cos(\bm{x}_{+k}, \bm{x}_i)\right),\\
    &= 2\eta(A_+-A_-),
\end{aligned}
\label{eq:l1_mask_2}
\end{equation}

where

\begin{equation}
\begin{aligned}
    A_+ &= \sum_{k=1}^N \|\bm{h}_{+k}\odot \bm{m}_i\|_1 cos(\bm{x}_{+k}, \bm{x}_i),\\
    A_- &= \sum_{k=1}^N \|\bm{h}_{-k}\odot \bm{m}_i\|_1 cos(\bm{x}_{-k}, \bm{x}_i).
\end{aligned}   
\end{equation}

~

\textbf{Change of the }$\bm{l_2}$\textbf{ norm:}

Recall the local Taylor expansion gives us (\ref{eq:l2_taylor}):

\begin{equation}
    \Delta_{l_2} = (\bm{h}_i'-\bm{h}_i)^T\frac{1}{\|\bm{h}_i\|_2} \bm{h}_i + \mathcal{O}(\frac{1}{n^2}).
\end{equation}

Bring the $(\bm{h}_i'-\bm{h}_i)$ in (\ref{eq:h_change_2}) into the above equation, we get:

\begin{equation}
\begin{aligned}
    \Delta_{l_2} &= \left(-2\eta\sum_{k=1}^N \left(\bm{h}_{-k} \bm{x}_{-k}^T\bm{x}_i\odot \bm{m}_i - \bm{h}_{+k} \bm{x}_{+k}^T\bm{x}_i\odot \bm{m}_i\right)\right)^T\frac{1}{\|\bm{h}_i\|_2} \bm{h}_i + \mathcal{O}(\frac{1}{n^2}),\\
    &=-2\eta \sum_{k=1}^N\Bigg(\|\bm{h}_{-k}\odot \bm{m}_i\|_2 cos(\bm{x}_{-k}, \bm{x}_i) cos(\bm{h}_{-k}\odot \bm{m}_i, \bm{h}_i) \\
    &- \|\bm{h}_{+k}\odot \bm{m}_i\|_2 cos(\bm{x}_{+k}, \bm{x}_i) cos(\bm{h}_{+k}\odot \bm{m}_i, \bm{h}_i) \Bigg) + \mathcal{O}(\frac{1}{n^2}),\\
    &= 2\eta (B_+-B_-) + \mathcal{O}(\frac{1}{n^2}),
    \label{eq:dl2_r_2}
\end{aligned}
\end{equation}

where

\begin{equation}
\begin{aligned}
    B_+ &= \sum_{k=1}^N \|\bm{h}_{+k}\odot \bm{m}_i\|_2 cos(\bm{x}_{+k}, \bm{x}_i) cos(\bm{h}_{+k}\odot \bm{m}_i, \bm{h}_i),\\
    B_- &= \sum_{k=1}^N \|\bm{h}_{-k}\odot \bm{m}_i\|_2 cos(\bm{x}_{-k}, \bm{x}_i) cos(\bm{h}_{-k}\odot \bm{m}_i, \bm{h}_i).\\
\end{aligned}
\end{equation}

Now we have the change on both $l_1$ and $l_2$ norms. Next, we will bring them back to the sparsity comparison.

~

\textbf{Final step}

Recall comparing the value of the sparsity measures $S(\bm{h}_i)$ and $S(\bm{h}_i')$:

\begin{equation}
    \text{(Left~hand~side)~~~}\frac{\sqrt{n} - \frac{\|\bm{h}_i\|_1}{\|\bm{h}_i\|_2}}{\sqrt{n} - 1} \text{~~~v.s.~~~} \frac{\sqrt{n} - \frac{\|\bm{h}_i'\|_1}{\|\bm{h}_i'\|_2}}{\sqrt{n} - 1}\text{~~~(Right~hand~side)},
\end{equation}

is equivalent to compare

\begin{equation}
    \text{(Left~hand~side)~~~} \Delta_{l_1}\|\bm{h}_i\|_2 \text{~~~v.s.~~~}\|\bm{h}_i\|_1\Delta_{l_2}\text{~~~(Right~hand~side)}.
    \label{eq:compare_trans}
\end{equation}

Bringing the calculated $\Delta_{l_1}$ in (\ref{eq:l1_mask_2}) and $\Delta_{l_1}$ in (\ref{eq:dl2_r_2}) into the above comparison, we get:

\begin{equation}
    2\eta(A_+-A_-) \|\bm{h}_i\|_2 \text{~~~v.s.~~~}\|\bm{h}_i\|_1 \left(2\eta(B_+-B_-)+\mathcal{O}(\frac{1}{n^2})\right).
\end{equation}

Next, we devide both sides by $2\eta$. Since $\eta=\mathcal{O}(\frac{1}{n})$, dividing $\mathcal{O}(\frac{1}{n^2})$ by $2\eta$ results in $\mathcal{O}(\frac{1}{n})$:

\begin{equation}
     \|\bm{h}_i\|_2 (A_+-A_-) \text{~~~v.s.~~~}\|\bm{h}_i\|_1 \left((B_+-B_-)+\mathcal{O}(\frac{1}{n})\right).
\end{equation}

We can omit $\mathcal{O}(\frac{1}{n})$ as $n\rightarrow+\infty$. 

In conclusion, when requiring goodness decrease for negative samples and goodness increase for positive samples, asking an activation $\bm{h}_i$ becomes sparser after weight update:

\begin{equation}
    \text{(Left~hand~side)~~~}\frac{\sqrt{n} - \frac{\|\bm{h}_i\|_1}{\|\bm{h}_i\|_2}}{\sqrt{n} - 1} < \frac{\sqrt{n} - \frac{\|\bm{h}_i'\|_1}{\|\bm{h}_i'\|_2}}{\sqrt{n} - 1}\text{~~~(Right~hand~side)},
\end{equation}

is equivalent to requiring:

\begin{equation}
     \|\bm{h}_i\|_2 (A_+-A_-) < \|\bm{h}_i\|_1 (B_+-B_-),
\end{equation}

where

\begin{equation}
    \begin{aligned}
        A_+ & = \sum_{k=1}^N \|\bm{h}_{+k} \odot \bm{m}_i\|_1 cos(\bm{x}_{+k},\bm{x}_i),\\
        A_- & = \sum_{k=1}^N \|\bm{h}_{-k} \odot \bm{m}_i\|_1 cos(\bm{x}_{-k},\bm{x}_i),\\
        B_+ & = \sum_{k=1}^N\|\bm{h}_{+k}\odot \bm{m}_i\|_2 cos(\bm{h}_i, \bm{h}_{+k}\odot \bm{m}_i) cos(\bm{x}_{+k},\bm{x}_i),\\
        B_- & = \sum_{k=1}^N\|\bm{h}_{-k}\odot \bm{m}_i\|_2 cos(\bm{h}_i, \bm{h}_{-k}\odot \bm{m}_i) cos(\bm{x}_{-k},\bm{x}_i).
    \end{aligned}
\end{equation}

\begin{flushright}
\textbf{[Q.E.D.]}
\end{flushright}

\newpage
\section{Preliminary Experimental Results}
\subsection{Theorem \ref{theorem:1}}

~

\textbf{MNIST on a randomly initialized linear layer:}

Here we test how a real-world dataset satisfies the derived property in Theorem \ref{theorem:1}:

\begin{equation}
    \text{(Left~hand~side)~~~}\frac{\|\bm{h}_i\|_1}{\|\bm{h}_i\|_2} < \frac{\sum_{k=1}^N cos(\bm{x}_k,\bm{x}_i) \| \bm{h}_{k\bm{m}_i}\|_1}{\sum_{k=1}^N \|\bm{h}_{k\bm{m}_i}\|_2 cos(\bm{h}_{k\bm{m}_i}, \bm{h}_i) cos(\bm{x}_k,\bm{x}_i)} \text{~~~(Right~hand~side)}.
\end{equation}

We define a linear layer with ReLU activation initialized by $\mathbf{W}\sim\mathcal{N}(0,\frac{2}{n})$, and feed in the normalized MNIST images $\mathcal{X}=\{\bm{x}_1, \bm{x}_2, \cdots, \bm{x}_N\}$ with batch size $=N$. The corresponding outputs are $\mathcal{H}=\{\bm{h}_1, \bm{h}_2, \cdots, \bm{h}_N\}$. For each image, we check its values on the left hand side (LHS) and the right hand side (RHS) of Theorem \ref{theorem:1}. The results after traversing the whole training set is shown in the figure below:

\begin{figure}[h]
    \centering
    \includegraphics[width=0.99\linewidth]{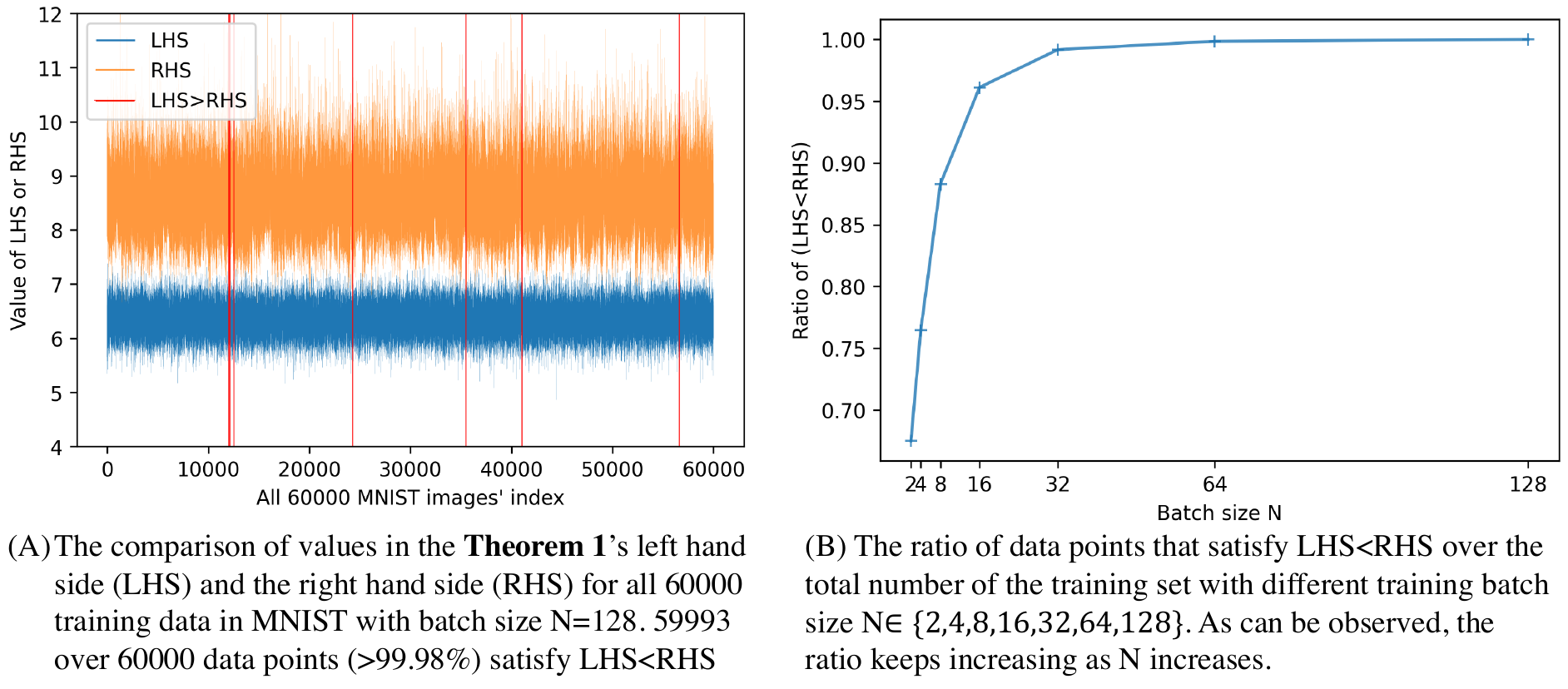}
    \caption{(A) With a randomly initialized linear layer that has n=128 neurons, we feed in MNIST images $\bm{x}$ to get their activations $\bm{h}$. Then, for each sample, we bring $\bm{x}$ and $\bm{h}$ into the equation in Theorem \ref{theorem:1} and visualized the values of LHS (blue) and RHS (orange). Almost all images (>99.98\%) in the MNIST dataset satisfies LHS<RHS. (B) Keeping all other hyper-parameters fixed, the larger the batch size, the higher ratio of input data points satisfy the derived property.}
    \label{fig:sparsity}
\end{figure}

Therefore, the derived property is satisfied in most of the cases. This is not surprising. Because decreasing goodness means decreasing all activation's $l_2$ norm, which would naturally lead many dimensions reduced to zero with the ReLU activation. Yet as training proceed, many input samples' activations would be totally silenced, making further sparser impossible. Therefore we may observe an decreasing in the ratio of data samples that become sparser as training proceed. This is tested in the next experiment:

~
\newpage
\textbf{MNIST while training}

As training proceed, it becomes more and more challenge to push the activation sparser. Which means the ratio of data points in a batch that satisfies Theorem \ref{theorem:1} shall keep decreasing as well.

Here, we train the linear layer on the MNIST dataset for two epochs (938 iterations with batch size N=128) to decrease the goodness of each batch. We plot both the ratio of data points that become sparser over the batch size N in each iteration after the weight update, and the ratio of data points that satisfy Theorem \ref{theorem:1} among the whole batch. The result is shown in the figure below:

\begin{figure}[h]
    \centering
    \includegraphics[width=0.7\linewidth]{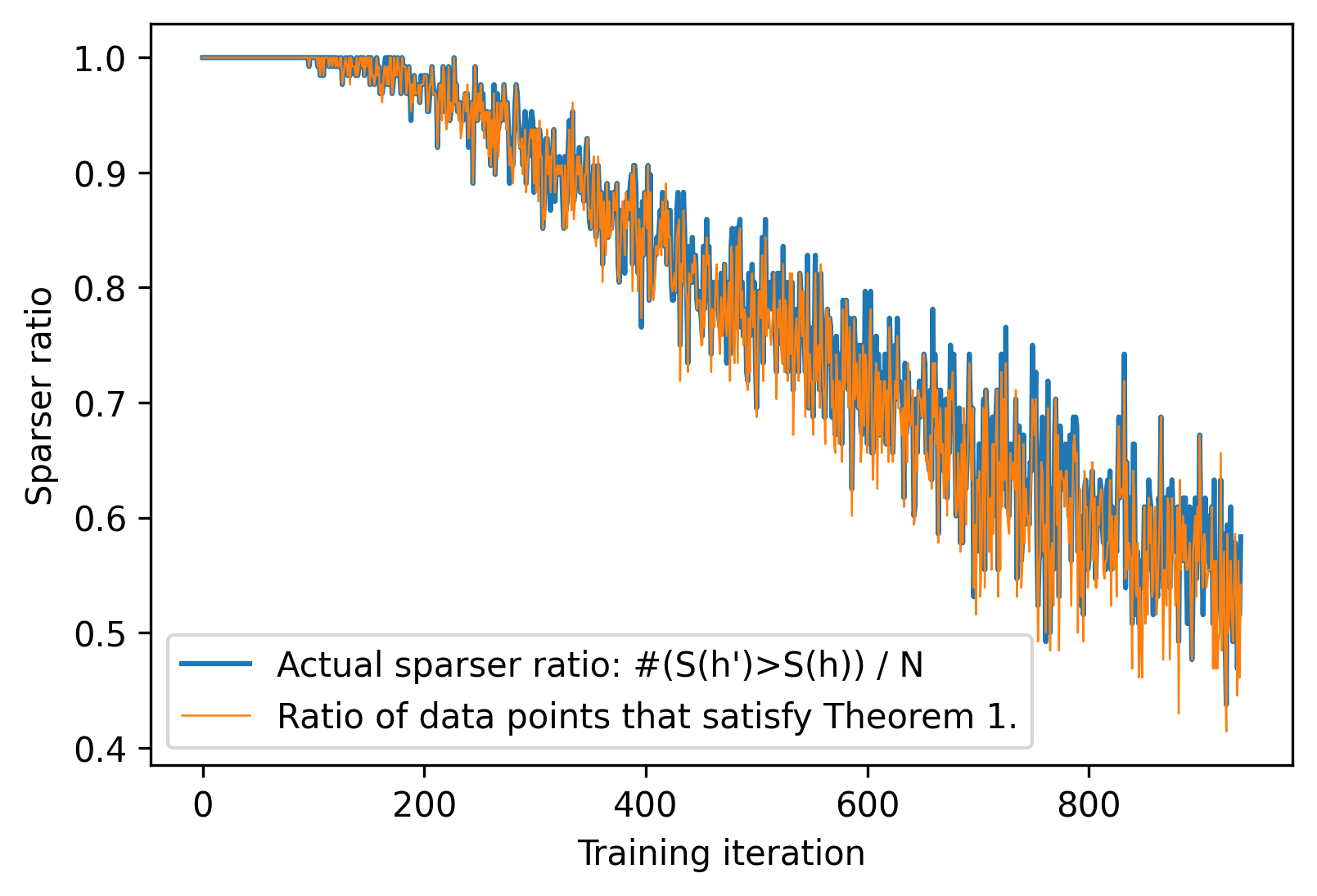}
    \caption{The blue line keeps track of the actual ratio of data points that become sparser over the batch size N in each iteration after the weight update. The orange line is the ratio of data points that satisfy Theorem \ref{theorem:1} in each iteration. (Loss is defined as the square sum of the activation. Hyper-parameters: Batch size N=128, neuron number n=2000, the optimizer adopted is the stochastic gradient descent (SGD) with learning rate = 0.001, momentum=0).}
    \label{fig:sparsity_ratio}
\end{figure}

One can observe that the two lines follow the same decreasing trend as training proceed. Although the two lines do not perfectly overlap, the result still demonstrated that the proposed Theorem is a good indicator about the change of sparseness.

\newpage
\subsection{Theorem \ref{theorem:2}}

We run the single layer forward forward algorithm (FFA) on MNIST with positive examples and negative examples constructed the same way as in the original FFA paper: for the positive samples, the inputs are the combination of both the images and their corresponding labels, while the negative samples' inputs are the combination of images with randomly generated wrong labels.

We keep track of the ratio in each batch that becomes sparser for both positive and negative samples, together with the ratio in a batch that satisfies Theorem \ref{theorem:2}. The result is shown below:

\begin{figure}[h]
    \centering
    \includegraphics[width=0.99\linewidth]{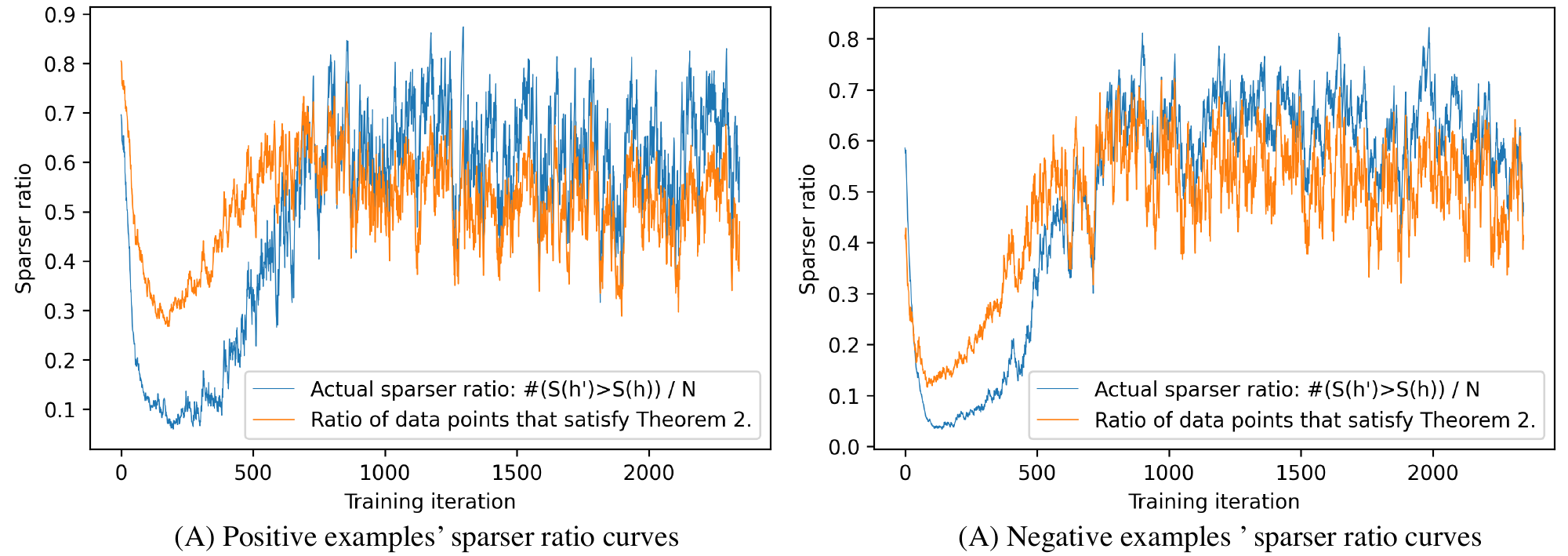}
    \caption{In both (A) and (B), the blue lines keep track of the actual ratio of data points that become sparser over the batch size N in each iteration after the weight update. The orange lines are the ratios of data points that satisfy Theorem \ref{theorem:2} in each iteration. (A) shows the two ratios for the positive data samples, (B) shows the two ratios of the negative data samples.  (We trained on MNIST for 5 epochs, with batch size N=128 for both the positive and negative samples, neuron number n=2000, the optimizer is SGD with learning rate = 0.001, momentum=0.)}
    \label{fig:sparsityB}
\end{figure}

From the result, we observe: 1) The proposed Theorem \ref{theorem:2} follows the same trend as the actual result, which shows the correctness of the derived theorem. The mismatch  may need further investigation. 2) The sparser ratio is larger then 50\% for most of the time for both positive and negative samples, which support the experimental findings that both the positive and negative samples would have sparser activations as trained by the forward forward algorithm.

\printbibliography


\end{document}